# DETECTION OF SURFACE DEFECTS ON CERAMIC TILES BASED ON MORPHOLOGICAL TECHNIQUES

*Grasha Jacob[1], R. Shenbagavalli[2], S. Karthika[3]*
[1] Associate Professor, [2] Assistant Professor, [3] Research Scholar
Department of Computer Science, Rani Anna Government College for Women, Tirunelveli
e-mail: arthisha2015@gmail.com

**ABSTRACT**

*Ceramic tiles have become very popular and are used in the flooring of offices and shopping malls. As testing the quality of tiles manually in a highly polluted environment in the manufacturing industry is a labor-intensive and time consuming process, analysis is carried out on the tile images. This paper discusses an automated system to detect the defects on the surface of ceramic tiles based on dilation, erosion, SMEE and boundary detection techniques.*

*Keywords: Ceramic Tiles, Tile Defect, Structuring Element, Dilation, Erosion, Quality Control*

## 1.0 INTRODUCTION

The techniques of image processing are being used in in the Quality Control departments of Glass industry, Textile industry, and Ceramic industry. In ceramic tile industry many automated systems have been developed to analyze the quality of tiles. Generally these automated systems have been developed to detect the defects only for plane surface tiles as it is very difficult to detect the defects on designed tiles. Defects on designed tiles can be identified based on morphological techniques. Morphological image processing is a set of non-linear operations associated with the shape or morphology of features in an image. Montreal & Quebec, (2006) described the need for detecting edges of images and proposed edge detection methods such as Canny and Sobel [1]. These methods were applied on images of ceramic tiles with defect based on color and surface. Ar & Akgul, (2008) presented marble tile feature extraction system which can be easily used for any classification system [2]. Image processing techniques especially Gabor filtering was engaged to differentiate between different marble textures and a new verification method on the basis of the inter-expert variability was presented and the percentages of veins, spots, and swirls on the marble images were calculated. Therefore, the system was considered as the core engine of a very portable marble tile classification system. Hocenski & Vasilic, (2006) proposed the method for faults detection based on edge detection techniques using Canny edge detector [3]. Histogram subtraction method was used to fix problem of defining hysteresis threshold and edge and surface faults were identified.

Morphological techniques scan an image with a small shape or template called a **structuring element**. The structuring element serves as a key in identifying the defect on tiles. The structuring element is positioned at all possible locations in the image and it is compared with the corresponding neighborhood of pixels. Certain operations test whether the element fits within the neighborhood and others test whether it hits or intersects the neighborhood.

Table 1. Different types of Defects on Tiles

| *Defect* | *Description* |
|---|---|
| Blob | Water drop spot on the surface |
| Corner | Break down in the corner of tile |
| Crack | Break in tile |
| Edge | Break in edge |
| Glaze | Blurred surface on the tile |
| Pinhole | Isolated black-white pinpoint spot |
| Scratch | Scratch on surface |
| Spot | Discontinuity of color on the surface |

The morphological operations like Dilation, Erosion, Dilation and Erosion, Simple Morphological Edge Extraction (SMEE), and Boundary Extraction techniques are used in identifying the different types of defects. This paper proposes an efficient defect detection and classification technique that would find out the defects on ceramic tiles at a higher rate within a very short span of time. Table 1 describes the different types of defects on the surface of ceramic tiles.

## 2.0 MORPHOLOGICAL OPERATORS AND OPERATIONS

The morphological operators centered on dilation and erosion are Opening and Closing. Opening smoothens the contour of images by breaking narrow gaps and eliminating small holes or thin protrusions. The Opening of an image A by a structuring element B (denoted by **A ° B**) is given by the successive operations of erosion and dilation and is given by the expression

$$A \circ B = (A \ominus B) \oplus B \quad \ldots \ldots \ldots (1)$$

Closing tends to smoothen the contour of an image by fusing narrow breaks and long thin gulfs, and eliminates small holes by filling gaps in the contour. The Closing of an image A by a structuring element B (denoted by A·B) is given by the successive operations of dilation and erosion and is represented by the expression

$$A \cdot B = (A \oplus B) \ominus B \quad \ldots \ldots \ldots (2)$$

Dilation and erosion are the fundamental morphological operations. Dilation adds pixels to the boundaries of objects in an image, while erosion removes pixels on object boundaries. The number of pixels added or removed from the objects in an image depends on the size and shape of the structuring element used to process the image.

## 2.1 DILATION

Dilation is the addition of a pixel at object boundary based on a structuring element. It is defined as the maximum value in the window. The image after dilation will be brighter with an increase in intensity. It enlarges the image objects by changing pixels with the value of "0" to "1". Foreground pixels are denoted by 1's and background pixels by 0's. To compute the dilation of a binary input image based on the structuring element consider each of the background pixels in the input image and superimpose the structuring element on top of the input image so that the origin of the structuring element coincides with the input pixel position. If anyone pixel in the structuring element coincides with a foreground pixel in the mage underneath, then the input pixel is set to the foreground value. If all the corresponding pixels in the image are background, however, the input pixel is left at the background value. The dilation operation is pictorially epitomized in Figure 1.

Dilation operation can be applied using the formula:

$$\{[(I \bullet Se) \circ Se] \bullet Se\} \oplus Se - \{[(I \bullet Se) \circ Se] \bullet Se\} \quad \ldots \ldots \ldots (3)$$

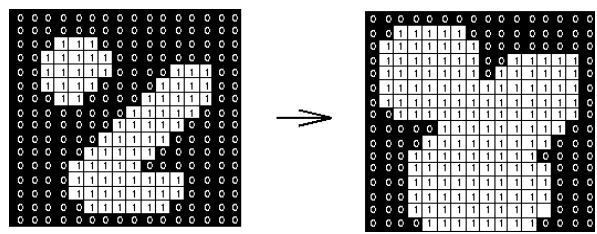

**Figure 1:** Dilation Operation

### 2.1.1 PROCEDURE FOR DILATION

**Step 1:** Obtain the input image.

**Step 2:** Convert the images into binary image.

**Step 3:** Apply closing operator on the binary image using structuring element.

**Step 4:** Apply opening operator on the resulting image obtained in step 3 with the same structuring element.

**Step 5:** Apply closing operator once again on the resultant image obtained in step 4.

**Step 6:** Make a copy of the closed image obtained in Step 5.

**Step 7:** Dilate the image obtained in Step 5 with the same structuring element.

**Step 8:** Finally, subtract the closed image in Step 6 from the dilated image obtained in Step 7.

**Step 9:** Compute the pixel count of the image obtained in Step 8.

### 2.2. EROSION

The inverse of dilation is erosion and it is defined as the minimum value in the window. It removes the pixel from the object boundary based on the structuring element. The image after erosion will be darker than the original image. Erosion shrinks the images by altering the pixels with the value of "1" to "0". The erosion operation is pictorially exemplified in Figure 2.

Erosion operation can be applied using the formula:
$$\{[(I \bullet Se) \circ Se] \bullet Se\} \ominus Se + \{[(I \bullet Se) \circ Se] \bullet Se\} \quad \ldots \ldots \ldots (4)$$

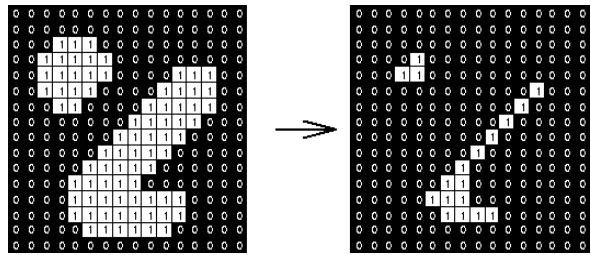

**Figure 2:** Erosion Operation

#### 2.2.1 PROCEDURE FOR EROSION

**Step 1:** Obtain the input image.

**Step 2:** Convert the image into binary image.

**Step 3:** Apply closing operator on the binary image using structuring element.

**Step 4:** Apply opening operator on the image obtained in step 3 with the same structuring element.

**Step 5:** Apply closing operator on the resultant image obtained in step 4.

**Step 6:** Make a copy of the closed image obtained in Step 5.

**Step 7:** Erode the image obtained in Step 5 with the same structuring element.

**Step 8:** Finally, add the closed image obtained in Step 6 with the eroded image obtained in Step 7.

**Step 9:** Compute the pixel count of the image obtained in Step 8.

### 2.3. SIMPLE MORPHOLOGICAL EDGE EXTRACTION (SMEE)

In Simple Morphological Edge Extraction, the query image is initially dilated and then the original image is subtracted from the dilated image.

SMEE is represented by the expression:

$$\boxed{SMEE = (A \oplus B) - A}$$

$$\ldots \ldots \ldots (5)$$

### 2.3.1 PROCEDURE FOR SMEE

**Step 1:** Obtain the input image.

**Step 2:** Dilate the input image based on the structuring element.

$$D1 = imdilate\ (img,\ se) \qquad \ldots\ldots\ldots (6)$$

**Step 3:** Subtract the input image from the dilated image.

$$SMEE = D1 - img \qquad \ldots\ldots\ldots (7)$$

**Step 4:** Compute the pixel count of the image obtained in Step 3.

### 2.4. BOUNDARY EXTRACTION

The boundary of a set A, denoted as β (A), can be obtained by first eroding A by B and then applying the set difference operation between A and its erosion.

The boundary extraction operation is pictorially symbolized in Figure 3.

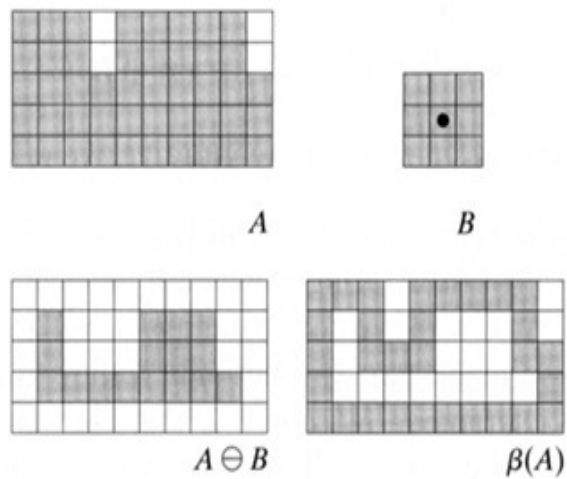

**Figure 3:** Boundary Extraction

Boundary Extraction can be denoted as follows:

$$\beta\ (A) = A - (A\ \Theta\ B) \qquad \ldots\ldots\ldots (8)$$

### 2.4.1 PROCEDURE FOR BOUNDARY EXTRACTION

**Step 1:** Obtain the input image.

**Step 2:** Erosion operation is applied on the input image, which increases on the black pixels.

$$E1 = imerode\ (img,\ se) \qquad \ldots\ldots\ldots (9)$$

**Step 3:** Subtract the input image from the eroded image.

$$BE = img - E1 \qquad \ldots\ldots\ldots (10)$$

**Step 4:** Compute the pixel count of the image obtained in Step 3.

### 3.0 METHODOLOGY

To identify the different types of defects on the surface of tiles based on the above said procedures the morphological operations - Dilation, Erosion, Dilation and Erosion, Simple Morphological Edge Extraction (SMEE), and Boundary Extraction techniques are used. On the basis of the pixel count of both the reference and test images, the ceramic tile images are tested and classified as either defect-free or defective. The pixel count of the image is computed by determining the number of picture elements (pixels).

In each method, the pixel count of the test image and that of the standard defect free image (reference image) are calculated. If the pixel count of the tile image that is tested for defect ($D_1$) is greater than the pixel count of the reference image (defect free image) ($R_1$), then it is classified as a defective tile otherwise it is a defect-free tile.

The defect, $\Delta d$ of the image is given by the relation

$$\Delta d = R_1 - D_1 \qquad \ldots \ldots \ldots (11)$$

where $R_1$ is the pixel count of the standard defect free image (reference image) and $D_1$ is the pixel count of the image being tested. For a defective tile, the defect, $\Delta d$ of the image will be negative ($< 0$).

The PSNR and MSE values are also calculated for each of the method applied to obtain a more precise and clear idea of the defective images. In addition, comparison is done based on the time complexity to identify the method that finds out the different types of defects quickly.

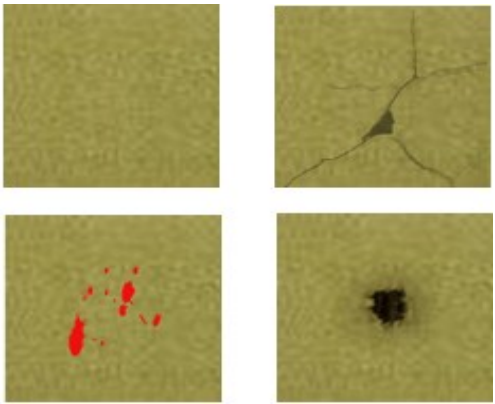

**Figure 4:** Reference image and Defective images

## 4. EXPERIMENTAL RESULTS AND ANALYSIS

In this work, in order to detect and identify the different types of defects such as Crack, Pinhole, Blob and Spot, two hundred ceramic tile images were considered. Figure 4 represents the reference tile image (Standard defect-free image) and three defective tile images. Figure 5 shows the implementation of dilation operation on the three defective tile images.

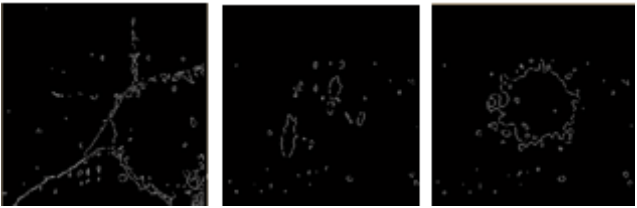

**Figure 5:** Dilation

Figure 6 shows the implementation of erosion operation on the three defective tile images.

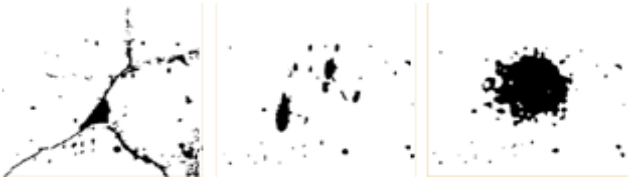

**Figure 6:** Erosion

Figure 7 shows the implementation of SMEE operation on the three defective tile images.

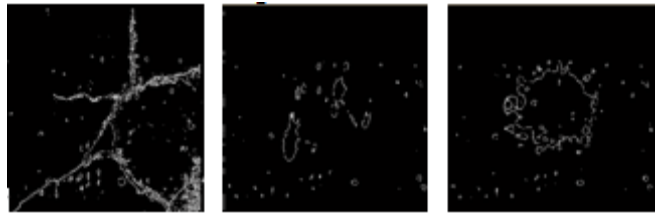

**Figure 7:** SMEE

Figure 8 shows the implementation of boundary extraction operation on the three defective tile images.

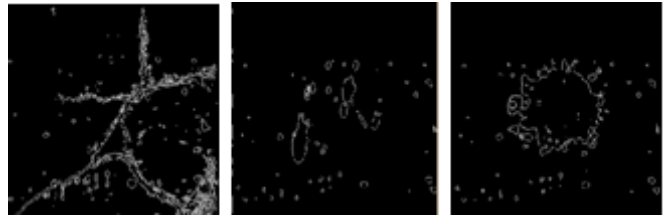

**Figure 8:** Boundary Extraction

Table 2 tabulates the pixel count of the standard defect free tile image (reference tile) and that of the defective tile images that are tested based on the four different operations.

**Table 2:** Pixel Count of the tile images

| Method | Defect free Pixel count | Pixel count of Defect | | |
|---|---|---|---|---|
| | | Crack | Paint Conceals | Spot |
| **Dilation** | 315 | 2904 | 1532 | 1029 |
| **Erosion** | 425 | 1426 | 1670 | 1185 |
| **SMEE** | 510 | 6823 | 2169 | 1501 |
| **Boundary Extraction** | 704 | 7964 | 2416 | 1916 |

It is clear from Table 2 that the pixel count of the defective tiles are greater than that of the standard defect free tile (reference tile image) indicating that the tested tile images are defective.

Peak signal-to-noise ratio, often abbreviated as PSNR gives an approximation to human perception of reconstruction quality. A higher PSNR value generally indicates that the reconstruction is of higher quality.

$$PSNR = 10 \cdot \log_{10}\left(\frac{MAX_I^2}{MSE}\right)$$

… … … (12)

where $MAX_I$ is the maximum possible pixel value of the image. $MAX_I$ is 255When the pixels are represented using 8 bits per sample.

Table 3 tabulates the PSNR value obtained for the four different operations on the three defective tiles.

**Table 3:** PSNR

| Defect Type | Dilation | Erosion | SMEE | Boundary Extraction |
|---|---|---|---|---|
| Crack | 17.0101 | 12.4384 | 13.4492 | 12.7368 |
| Paint Conceals | 20.9963 | 15.5265 | 19.2789 | 18.178 |
| Spot | 20.7864 | 9.5647 | 19.3205 | 18.9877 |

Given a noise-free m x n monochrome image I and its noisy approximation K, the mean squared error (MSE) is defined as:

$$MSE = \frac{1}{m\,n} \sum_{i=0}^{m-1} \sum_{j=0}^{n-1} [I(i,j) - K(i,j)]^2$$

… … … (13)

Table 4 tabulates the PSNR value obtained for the four different operations on the three defective tiles.

**Table 4:** MSE

| Defect Type | Dilation | Erosion | SMEE | Boundary Extraction |
|---|---|---|---|---|
| Crack | 0.011991 | 0.05703 | 0.04519 | 0.05325 |
| Paint Conceals | 0.00795 | 0.02801 | 0.01180 | 0.01521 |
| Spot | 0.00834 | 0.11005 | 0.01169 | 0.01262 |

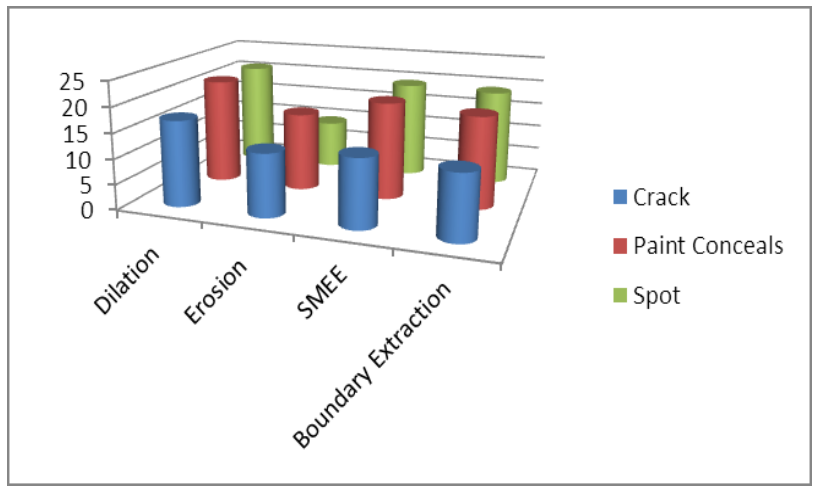

**Figure 9:** Graphical Representation of PSNR

Figure 9 displays the graphical representation of PSNR values and Figure 10 displays the graphical representation of MSE values. From Table 3 and Table 4, PSNR and MSE vales tabulated for the defective tiles help to decide the method that gives a clear vision of the defective portion on the tile. It is clear that a high PSNR value and a low MSE value will clearly identify the type of defect on the tiles.

Based on Table 3, PSNR value of dilation method is high and based on Table 4, MSE value of the same method (dilation) has a low value. Therefore, from the above four methods, dilation method is considered to be the best method to detect the different types of defects.

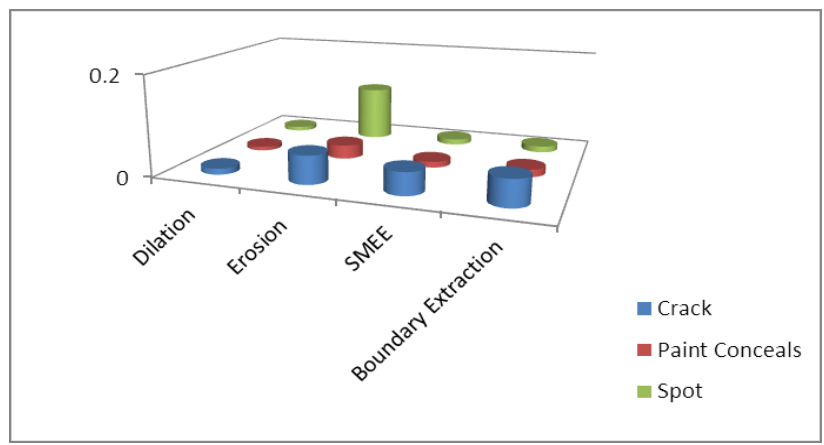

**Figure 10:** Graphical Representation of MSE

The time taken to detect the different defects according to different morphological methods is calculated. Table 5 represents the time complexity involved in the four different operations. According to Table 5, Simple Morphological Edge Extraction operation takes the least time to detect the cracks, Pinholes and Blob defects. Figure 11 represents the time taken to detect the different types of defects.

**Table 5:** Time Complexity

| Defect Type | Dilation (t0) | Erosion (t1) | SMEE (t2) | Boundary Extraction(t3) |
|---|---|---|---|---|
| Crack | 1.2776 | 1.2550 | 0.5482 | 0.5926 |
| Paint Conceals | 1.0248 | 1.4022 | 0.5596 | 0.5825 |
| Spot | 1.1441 | 1.4817 | 0.5399 | 0.6167 |

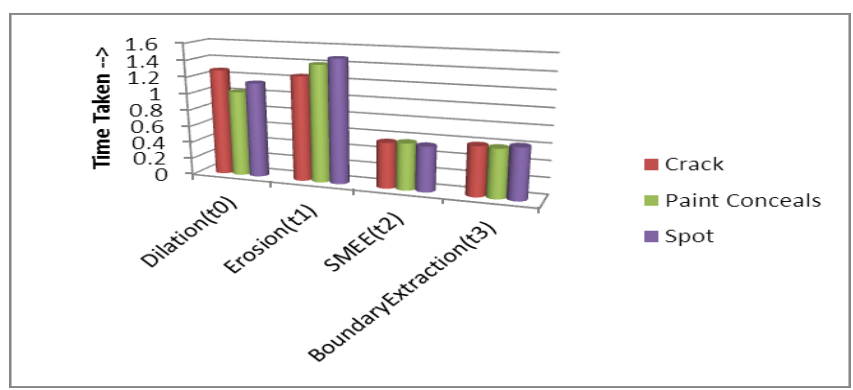

**Figure 11:** Graphical Representation of time complexity

Thus the experimental results clearly reveal that the dilation operation is well suited for identification of defects in ceramic tile images and SMEE is considered to be the best method based on time complexity.

## 6. CONCLUSION

In ceramic tile manufacturing industries, quality control is a labor intensive process and has to be performed in a highly polluted industrial environment. This work will assist the quality control department of tile manufacturing industries in determining and identifying the defects in the surface of the ceramic tiles based on the morphological operations such as Dilation, Erosion, SMEE and Boundary Extraction with the help of the tile images.